 \documentclass[pdflatex,sn-mathphys-num]{sn-jnl}

\usepackage{graphicx}%
\usepackage{multirow}%
\usepackage{amsmath,amssymb,amsfonts}%
\usepackage{amsthm}%
\usepackage{mathrsfs}%
\usepackage[title]{appendix}%
\usepackage{xcolor}%
\usepackage{textcomp}%
\usepackage{manyfoot}%
\usepackage{booktabs}%
\usepackage{algorithm}%
\usepackage{algorithmicx}%
\usepackage{algpseudocode}%
\usepackage{listings}%
\usepackage{threeparttable} 
\usepackage{pifont}
\usepackage{tikz-cd}
\usepackage{adjustbox}
\usepackage{xcolor}
\usepackage{colortbl}
\usepackage{threeparttable}

\theoremstyle{thmstyleone}%
\theoremstyle{thmstyletwo}%

\theoremstyle{thmstylethree}%

\begin{document}

\title[A Causality-Aware Spatiotemporal Model for Multi-Region and Multi-Pollutant Air Quality Forecasting]
{A Causality-Aware Spatiotemporal Model for Multi-Region and Multi-Pollutant Air Quality Forecasting}
 

\author[1,2]{\fnm{Junxin} \sur{Lu}}\email{junxinlu.ecnu@gmail.com}

\author*[1]{\fnm{Shiliang} \sur{Sun}}\email{slsun@sjtu.edu.cn}


\affil*[1]{\orgdiv{School of Automation and Intelligent Sensing}, \orgname{Shanghai Jiao Tong University}, \orgaddress{\street{800 Dongchuan Road}, \city{Shanghai}, \postcode{200240},  \country{China}}}

\affil[2]{\orgdiv{School of Computer Science and Technology}, \orgname{East China Normal University}, \orgaddress{ \city{Shanghai}, \postcode{200062},  \country{China}}}
%


\abstract{
Air pollution, a pressing global problem, threatens public health, environmental sustainability,
 and climate stability.    
Achieving accurate and scalable forecasting across spatially distributed monitoring stations
 is  challenging due to intricate multi-pollutant interactions, evolving meteorological conditions, and region-specific spatial heterogeneity.
To address this challenge,  we propose AirPCM, a novel deep spatiotemporal forecasting  model that 
 integrates multi-region, multi-pollutant dynamics with explicit meteorology-pollutant causality modeling.
 Unlike existing methods limited to single pollutants or localized regions,
  AirPCM  employs a unified architecture to jointly capture cross-station spatial correlations, temporal auto-correlations, 
and meteorology-pollutant dynamic causality.
This empowers fine-grained, interpretable multi-pollutant forecasting across 
varying geographic and temporal scales, including sudden pollution episodes.
Extensive evaluations on multi-scale real-world datasets demonstrate that AirPCM 
consistently surpasses state-of-the-art baselines in both predictive accuracy and generalization capability.
Moreover, the long-term forecasting capability of AirPCM provides actionable 
insights into future air quality trends and potential high-risk windows,
 offering timely support for evidence-based environmental governance and carbon mitigation planning.
}



\maketitle

\section{Introduction} 
 Air pollution is a pressing global problem, threatening human health, 
 impeding sustainable urban development, and exacerbating global climate change 
 \cite{li2019air,azimi2024unveiling,zhang2017transboundary}.
The World Health Organization reports that over
 90$\%$ of the global population is chronically exposed to pollutant levels exceeding safe thresholds, resulting
 in millions of premature deaths annually from respiratory and cardiovascular diseases 
 \cite{rentschler2023global,lelieveld2015contribution,brauer2012exposure}. 
In the face of escalating air quality challenges, developing high-precision, scalable air quality forecasting models 
is essential not only for daily health management and ecological governance but also for advancing global climate goals, 
such as carbon peaking and neutrality. 
While recent advances in multi-source environmental monitoring and deep learning have improved air quality predictions
\cite{subramaniam2022artificial,liu2022data,krupnova2022environmental}, 
existing methods face significant limitations in scalability, generalizability, and practical applicability.

\textbf{Limitations of single-region, single-pollutant forecasting paradigms.}
Most existing air quality forecasting models are constrained by a single-pollutant paradigm within specific regions,
such as PM$_{2.5}$  prediction in Beijing, China \cite{hettige2024airphynet,tian2025air,wang2020pm2}. 
This narrow forecasting paradigm fails to capture the dynamically coupled patterns of 
air quality across multiple regions and pollutants on a global scale.   
In reality,  the Air Quality Index (AQI) \cite{mendoza2021review} depends on the combined effects of multiple pollutants,
including PM$_{2.5}$, PM$_{10}$, carbon monoxide (CO), nitrogen dioxide (NO$_2$), 
sulfur dioxide (SO$_2$), and ozone (O$_3$) \cite{world2021global,akbarzadeh2018association}.
Effective policymaking and risk assessment require an integrated analysis of these pollutants.
Moreover, forecasting models tailored to specific regions often lack transferability across diverse economic and environmental
contexts, particularly in developing countries with severe pollution challenges \cite{wang2012air,vallero2025fundamentals}, 
leading to high deployment costs and limited global applicability.

\textbf{Scalability and generalizability constraints in modeling real-world air pollution dynamics.}
Current  air quality forecasting models, whether data-driven or physics-guided,
 face significant challenges in capturing the complex dynamics of real-world air pollution. 
Data-driven  methods \cite{guo2019attention,shang2021discrete,zhao2023multiple,wang2020pm2}  
excel at  identifying statistical patterns in historical data but often 
overlook underlying physical transport
 and inter-dependencies.
  This limits their interpretability and generalizability across diverse geographies 
 and meteorological conditions. 
Conversely, physics-guided models \cite{rubanova2019latent,chen2018neural,lechner2020learning,mohammadshirazi2023novel}  
 incorporate domain knowledge through governing physical equations (e.g., diffusion-advection \cite{hwang2021climate,choi2023climate,ji2022stden}
 or physical laws \cite{choi2023gread}) 
 and region-specific parameters, yielding promising results in localized settings. 
 However, their reliance on region-specific calibrations and assumptions of quasi-static systems restricts their
 applicability to multi-region and multi-pollutant forecasting.
 Furthermore, both methods struggle to adapt to sudden changes in air pollution dynamics, 
 such as those triggered by meteorological anomalies (e.g., thermal inversions or extreme wind) \cite{zhang2012real,sokhi2021advances}. 
 These limitations hinder their scalability for large-scale, multi-region, and multi-pollutant 
 forecasting tasks \cite{hettige2024airphynet,tian2025air}. 
 Therefore, it is crucial and develop  robust, interpretable, and transferable models, 
which can  more effectively inform global environmental policy 
and public health interventions across diverse spatial and temporal scales.

\textbf{Limited scale of benchmark datasets.}
Existing datasets are often restricted to monitoring within individual cities or countries \cite{wang2020pm2,10.1145/3219819,liang2023airformer},
lacking comprehensive environmental over multiple regions and pollutants. 
It is clear that strong performance on a single region and a single pollutant often fails to
identify globally transferable and robust models.
The limited scale of benchmark datasets thus significantly restricts
 their applicability in worldwide environmental governance and public health interventions.

To address these challenges, 
we propose \textbf{AirPCM}, a deep spatiotemporal model designed for 
accurate multi-region, multi-pollutant  \textbf{Air} quality forecasting though  \textbf{P}ollutant  \textbf{C}ausal  \textbf{M}odeling.
AirPCM makes the following multifaceted contributions.

\textbf{First}, AirPCM  simultaneously forecasts six major pollutants (PM$_{2.5}$, PM$_{10}$, O$_3$, NO$_2$, SO$_2$, and CO) 
across multi-region monitoring stations, 
enabling comprehensive AQI forecasting. 
This supports holistic environmental assessments and policy-making across diverse pollutant dimensions and geographic scales.

\textbf{Second}, AirPCM incorporates spatiotemporal and causal dependencies by jointly modeling  
  cross-station spatial correlations, temporal auto-correlations,
 and meteorology-pollutant dynamic causality.
This unified modeling strategy allows AirPCM to adapt to rapidly changing air conditions, 
improves forecasting robustness in complex environments, 
and enhances interpretability by explicitly revealing the driving meteorological factors behind pollution dynamics.

\textbf{Third}, to  fully support both short-term,  medium-term and long-term forecasting, 
we construct  two real-world benchmark datasets. The first, named as  
AirPCM-d, comprises daily records spanning from 2015 to 2025, 
covering over 170 major Chinese cities, 
and supports long-term air quality and AQI trend prediction.
The second dataset, AirPCM-h,  contains  hourly pollutant and meteorological 
observations collected from 2024 to 2025 covering over 430 monitoring stations across the
 United States, Europe, and China,   which can facilitate fine-grained, cross-regional forecasting and transferability evaluations. 
Extensive experiments on both datasets demonstrate that AirPCM consistently delivers superior 
multi-region, multi-pollutant forecasting performance, outperforming existing models in terms of accuracy, generalization, and robustness.

\begin{figure}[t]
\centering
\includegraphics[width=1.0\textwidth]{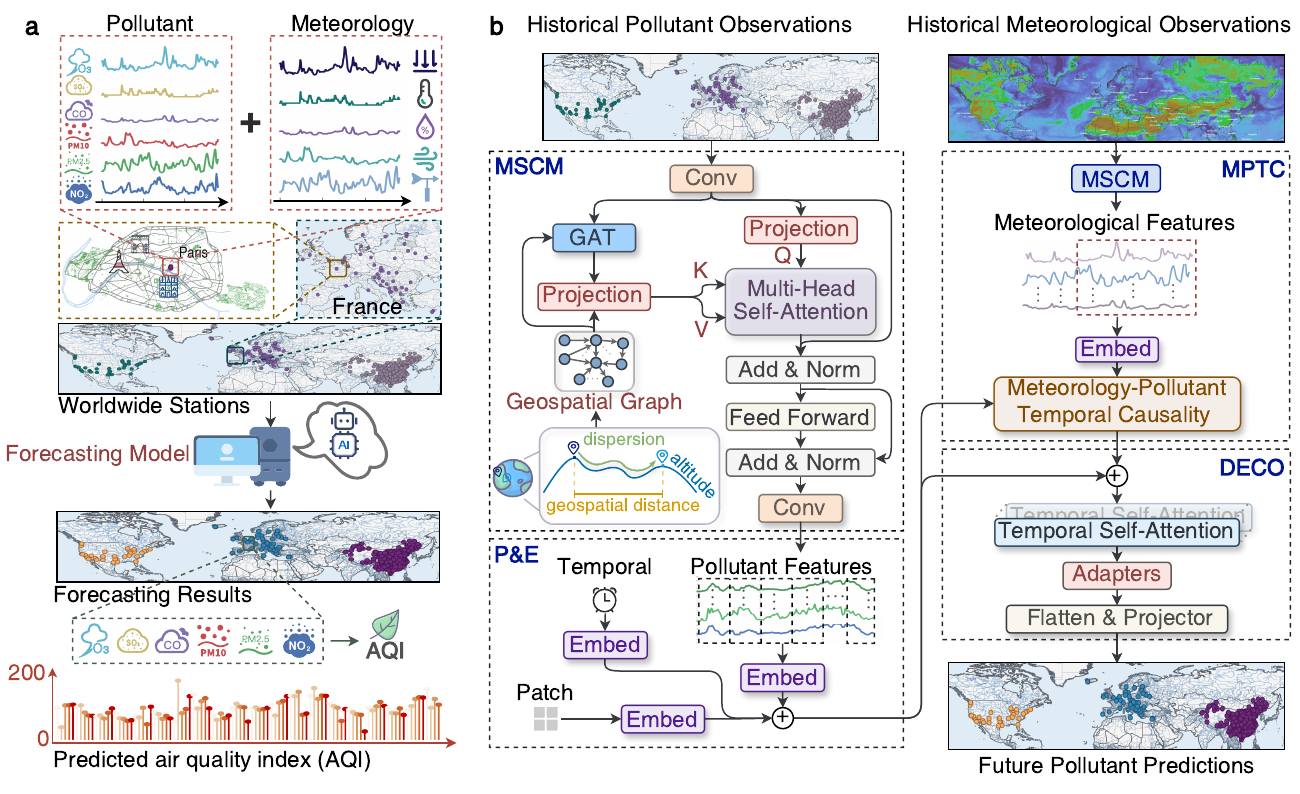}  
\caption{\textbf{Multi-region, multi-pollutant forecasting via a deep spatiotemporal model.} \textbf{a,}  Future pollutant concentrations and 
air quality indices  (AQI) are predicted from historical pollutant and meteorological observations collected at globally  distributed monitoring stations.
\textbf{b,}  The overall framework of AirPCM involves four key stages: 
multi-station
spatial correlation modeling (MSCM), followed by patching and embedding (P$\&$E),
meteorology-pollutant temporal causality modeling (MPTC) and finally decoding (DECO).
MSCM captures cross-station spatial correlations to support pollutant propagation across stations.
P$\&$E segments historical series into temporal patches and embeds them into latent spaces.
MPTC learns time-lagged meteorological-pollutant causal effect to model meteorology-driven pollutant dynamics.
Finally, DECO utilizes the encoded features to predict pollutant concentrations over the future time steps.
}
\label{overview}
\end{figure}  

\section{Results}
\subsection{Model overview}
Air quality forecasting is critical for public health, environmental protection, and sustainable urban development. 
Leveraging historical observations from air monitoring stations has become an effective approach to enable rapid and flexible air pollutant forecasting.
However, most existing methods are confined to single-region, single-pollutant prediction paradigms.
In reality, the global atmosphere is a highly coupled system characterized by complex interactions across regions and among multiple pollutants. 
Relying solely on historical data from a single area fails to capture the multifaceted spatial features of global air systems, 
while single-pollutant forecasting overlooks the inherently multi-pollutant nature of air quality analysis. 

This single-region, single-pollutant paradigm inherently ignores the collaborative, multi-region and multi-pollutant 
nature of air quality prediction. 
When extended to cross-regional and multi-pollutant scenarios, 
such methods result in trivial performance and incur prohibitive computational and storage costs, 
severely limiting their scalability to global applications.

To address these challenges, we develop AirPCM, a unified deep spatiotemporal 
model trained on historical observations from globally distributed air monitoring stations. 
This is motivated by the growing success of deep learning in air quality prediction
\cite{tian2025air,hettige2024airphynet,liang2023airformer,wang2020pm2,zhao2023multiple},  
to capturing the dynamic evolution of air pollution processes across spatial and temporal scales.
Our goal is to enable high-precision, wide-coverage forecasting of multiple pollutants across multiple regions, 
facilitating early warnings of air pollution events, days or even months in advance. 

The overall architecture of AirPCM is  illustrated in Fig.~\ref{overview},  
which aims to model the complex spatiotemporal and causal dependencies underlying multi-region, multi-pollutant forecasting.
We first focus on multi-station spatial correlation modeling (MSCM) to capture the local and global 
spatial interactions among worldwide distributed monitoring stations. 
Local spatial dependencies are captured using convolutional neural networks (CNNs) \cite{li2021survey}, 
while a geospatial graph is constructed based on station geographic coordinates to represent potential pathways of pollutant transport. 
To capture wider-range dependencies, we employ a graph attention network (GAT)  
\cite{velivckovic2018graph} to dynamically aggregate information from geographically adjacent 
stations, and a multi-head self-attention (MSA) mechanism \cite{vaswani2017attention} 
to model broader global spatial dependencies, 
ultimately yielding spatially contextualized representations for downstream temporal forecasting.
Then to capture the time-delay causal effects of meteorological variables on pollutant concentrations, 
we introduce a meteorology-pollutant temporal causality modeling module (MPTC). 
Specifically, by applying multi-head causal attention under temporal causal mask constraints,
we allow each pollutant patch to selectively attend  to relevant historical meteorological features within its causal window. 
This enables the model to learn interpretable temporal causal dependencies between meteorological variables and pollutants. 
Finally, we employ stacked temporal self-attention modules
 to capture intrinsic temporal dependencies within pollutant sequences. 
To accommodate the heterogeneous dynamics of different pollutants,
we integrate pollutant-specific adapters that adaptively modulate decoded features, 
thereby enhancing the forecasting precision of multiple pollutants.

In summary, by systematically uncovering the spatiotemporal and causal dependencies embedded in historical air quality data, 
AirPCM captures the evolving patterns of pollution dispersion and transformation, 
enabling accurate and interpretable multi-pollutant forecasting.

\begin{figure}[t]
\centering
\includegraphics[width=0.99\textwidth]{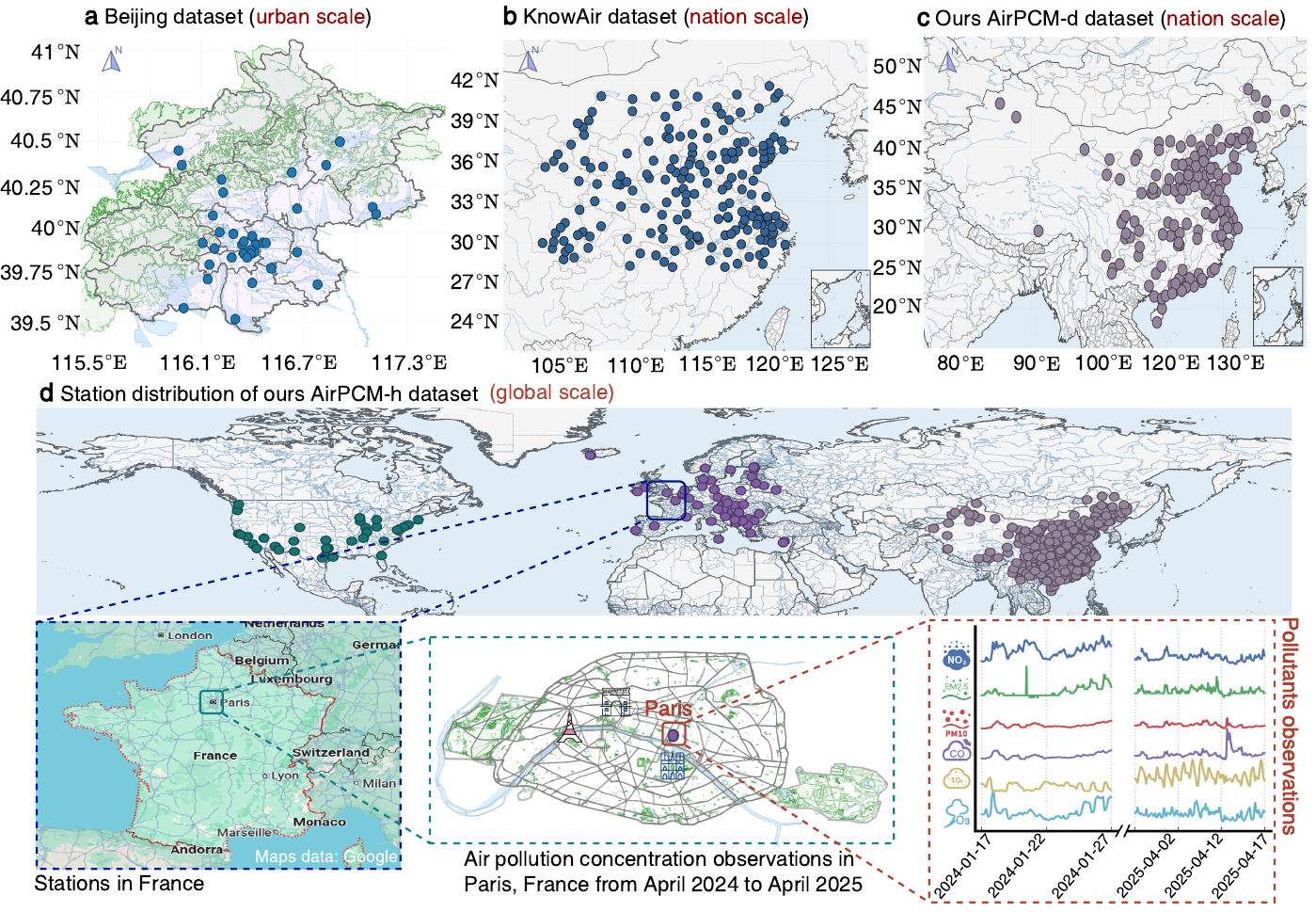}  
\caption{\textbf{Station distributions.}  
The urban-scale dataset \textbf{(a)} comprises 35 monitoring stations located in  
Beijing, China. 
The national-scale KnowAir dataset \textbf{(b)} encompasses measurements from 184 stations distributed across China.  
\textbf{c-d} depict our collected daily and hourly datasets.
The daily dataset \textbf{(c)}, AirPCM-d, comprises 
air pollution and meteorological records from 156 stations across China 
spanning the period 2015$-$2025. 
 The hourly dataset \textbf{(d)}, AirPCM-h,  includes  hourly  air pollution and meteorology data
from 17 January 2024 to 17 April 2025 across 453 monitoring stations, comprising  55 in Europe, 35 in the United States, and 363 in China.}
\label{mydata}
\end{figure}
  
\subsection{Forecasting performance of AirPCM}
We systematically evaluate the performance of AirPCM across four benchmarks, 
comprising two public datasets, Beijing and KnowAir, 
as well as two proprietary datasets, AirPCM-d and AirPCM-h (see Sec. \ref{dataset}). 
The station distributions of these datasets, as shown in Fig. \ref{mydata}, 
cover varying station densities and air pollution patterns from diverse regions.
AirPCM and all baseline models (see Sec. \ref{baseline_and_em}) 
are evaluated using Mean Absolute Error (MAE), Root Mean Square Error (RMSE), 
and Symmetric Mean Absolute Percentage Error (SMAPE) 
(see Sec. \ref{baseline_and_em}).

\textbf{Quantitative comparison.}  Table \ref{result} presents the PM$_{2.5}$  prediction  performance of AirPCM and the baselines on  
the Beijing and KnowAir datasets.  
The evaluation covers both the overall three-day forecasting accuracy and the ability to handle sudden changes. 
Because existing methods are primarily designed for single-region and single-pollutant prediction tasks, 
mostly targeting PM$_{2.5}$, we provide such a comparison focusing on PM$_{2.5}$ concentration forecasting.
All methods make predictions for the next three days based on the previous three days of observations, 
with input and prediction sequences each consisting of 24 steps at 3-hour intervals.
Following  Air-DualODE \cite{tian2025air} and Airformer \cite{liang2023airformer}, 
we further analyze prediction errors under sudden change scenarios. 
An sudden change is defined as a situation where the PM$_{2.5}$ concentration exceeds 75 $\mu g/m^3$ 
and fluctuates by more than ±20 $\mu g/m^3$ 
within the next three hours \cite{liang2023airformer}.

\begin{table*}[htbp]
\footnotesize
\centering
\setlength{\tabcolsep}{0.03cm} 
\caption{Overall forecasting performance comparison. 
Models predict the future 3 days from the past 3 days, with input and prediction sequences of length 24 (3-hour intervals). 
Bold indicates the best results; underlined denotes the second-best.}  
\begin{threeparttable}
\begin{tabular}{r|ccc|ccc|ccc|ccc}  \toprule
\rowcolor[HTML]{D8D6C2} & \multicolumn{6}{c|}{Beijing} & \multicolumn{6}{c}{KnowAir} \\  
\rowcolor[HTML]{D8D6C2}  & \multicolumn{3}{c|}{3 days} & \multicolumn{3}{c|}{Sudden Change} & \multicolumn{3}{c|}{3 days} & \multicolumn{3}{c}{Sudden Change} \\  
\rowcolor[HTML]{D8D6C2}  \multirow{-3}{*}{Methods\quad\quad}& MAE & RMSE & SMAPE & MAE & RMSE & SMAPE & MAE & RMSE & SMAPE & MAE & RMSE & SMAPE \\  \midrule
HA \cite{zhang2017deep}& 59.78 & 77.73 & 0.83 & 79.66 & 94.87 & 0.87 & 25.27 & 38.57 & 0.53 & 59.37 & 79.38 & 0.72 \\
\rowcolor[HTML]{EBEADE}  VAR \cite{toda1993vector} & 55.88 & 75.64 & 0.82 & 77.18 & 98.73 & 0.86 & 24.56 & 37.35 & 0.51 & 57.40 & 71.84 & 0.71 \\
Latent-ODE \cite{chen2018neural} & 43.35 & 63.75 & 0.79 & 69.71 & 92.67 & 0.77 & 19.99 & 30.85 & 0.46 & 42.59 & 57.90 & 0.50 \\
\rowcolor[HTML]{EBEADE}  ODE-RNN \cite{rubanova2019latent} & 43.60 & 63.67 & 0.79 & 70.22 & 93.12 & 0.78 & 20.57 & 31.26 & 0.45 & 42.07 & 57.76 & 0.51 \\
ODE-LSTM \cite{lechner2020learning} & 43.68 & 64.23 & 0.78 & 70.60 & 93.27 & 0.77 & 20.21 & 30.88 & 0.45 & 41.29 & 56.47 & 0.49 \\
\rowcolor[HTML]{EBEADE}  STGCN \cite{yu2018spatio} & 45.99 & 64.85 & 0.82 & 73.65 & 97.87 & 0.76 & 23.64 & 32.48 & 0.52 & 55.29 & 73.21 & 0.54 \\
DCRNN \cite{li2018diffusion} & 49.61 & 71.53 & 0.82 & 75.31 & 96.50 & 0.81 & 24.02 & 37.87 & 0.53 & 56.88 & 74.58 & 0.73 \\
\rowcolor[HTML]{EBEADE}  ASTGCN  \cite{guo2019attention} & 42.88 & 65.24 & 0.79 & 72.75 & 96.68 & 0.83 & 19.92 & 31.39 & 0.44 & 42.06 & 57.50 & 0.52 \\
GTS \cite{shang2021discrete} & 42.96 & 63.27 & 0.80 & 70.36 & 92.70 & 0.76 & 19.52 & 30.36 & 0.43 & 40.87 & 56.20 & 0.49 \\
\rowcolor[HTML]{EBEADE}  MTSF-DG \cite{zhao2023multiple} & 43.12 & 66.06 & 0.79 & 72.26 & 95.32 & 0.77 & 19.17 & 29.55 & 0.43 & 40.64 & 55.65 & 0.50 \\
PM25GNN \cite{wang2020pm2} & 45.23 & 66.43 & 0.82 & 72.45 & 95.34 & 0.78 & 19.32 & 30.12 & 0.43 & 40.43 & 55.49 & 0.49 \\
\rowcolor[HTML]{EBEADE}  Airformer \cite{liang2023airformer} & 42.74 & 63.11 & 0.78 & 68.80 & 91.16 & 0.75 & 19.17 & 30.19 & 0.43 & 39.99 & 55.35 & 0.49 \\
AirPhyNet \cite{hettige2024airphynet} & 42.72 & 64.58 & 0.78 & 70.03 & 94.60 & 0.78 & 21.31 & 31.77 & 0.47 & 43.23 & 58.79 & 0.50 \\
\rowcolor[HTML]{EBEADE}  Air-DualODE \cite{tian2025air} & \textbf{40.32} & \underline{62.04} & \underline{0.74} & \underline{66.40} & \underline{90.31} & \underline{0.73} & \underline{18.64} & \underline{29.37} & \underline{0.42} & \underline{39.79} & \underline{54.61} & \underline{0.49} \\ \midrule
\rowcolor[HTML]{D8D6C2} AirPCM\quad\quad & \underline{41.88} & \textbf{61.33} & \textbf{0.73} & \textbf{61.85} & \textbf{84.34} & \textbf{0.64} & \textbf{16.30} & \textbf{28.38} & \textbf{0.40} & \textbf{36.31} & \textbf{51.25} & \textbf{0.43}  \\
\rowcolor[HTML]{EBEADE}
Gain (\%) \quad 
& $-$3.9 & +1.1 & +1.4 
& +6.9 & +6.6 & +12.3
& +12.6 & +3.4 & +4.8
& +8.7 & +6.1 & +12.2 \\
\bottomrule
\end{tabular}
\begin{tablenotes}
\tiny
\item
\item $^*$ Gain=$\frac{\text{SOTA}-\text{AirPCM}}{\text{SOTA}}\times 100 \%$, where SOTA denotes the performance of the state-of-the-art baseline method.
\end{tablenotes}
\end{threeparttable}
\label{result}
\end{table*}

As shown in Table \ref{result}, AirPCM almost consistently outperforms all baselines 
across all datasets and evaluation scenarios.
 On the Beijing dataset, AirPCM achieves the lowest RMSE (61.33) and SMAPE (0.73) 
 for the 3-day prediction, 
 while also showing significant advantages under sudden changes by reducing MAE,  RMSE, and SMAPE by 6.9$\%$, 6.6$\%$, and 12.3$\%$, respectively,
 compared to the strongest baseline, Air-DualODE \cite{tian2025air}.
Similar improvements are observed again on the KnowAir dataset, 
where AirPCM lowers MAE to 16.30 for the regular 3-day forecast and to 36.31 under sudden changes, 
achieving relative gains of 12.6$\%$ (16.30 vs 18.64) and 8.7$\%$ (36.31 vs 39.79) over the 
prior state-of-the-art method. 
Furthermore, AirPCM provides up to 12.2$\%$ improvement (0.43 vs 0.49) in SMAPE, 
reflecting a better control capability over relative errors, 
which is particularly important for handling sudden changes in air quality.

\begin{figure*} 
\centering
\includegraphics[width=0.98\textwidth]{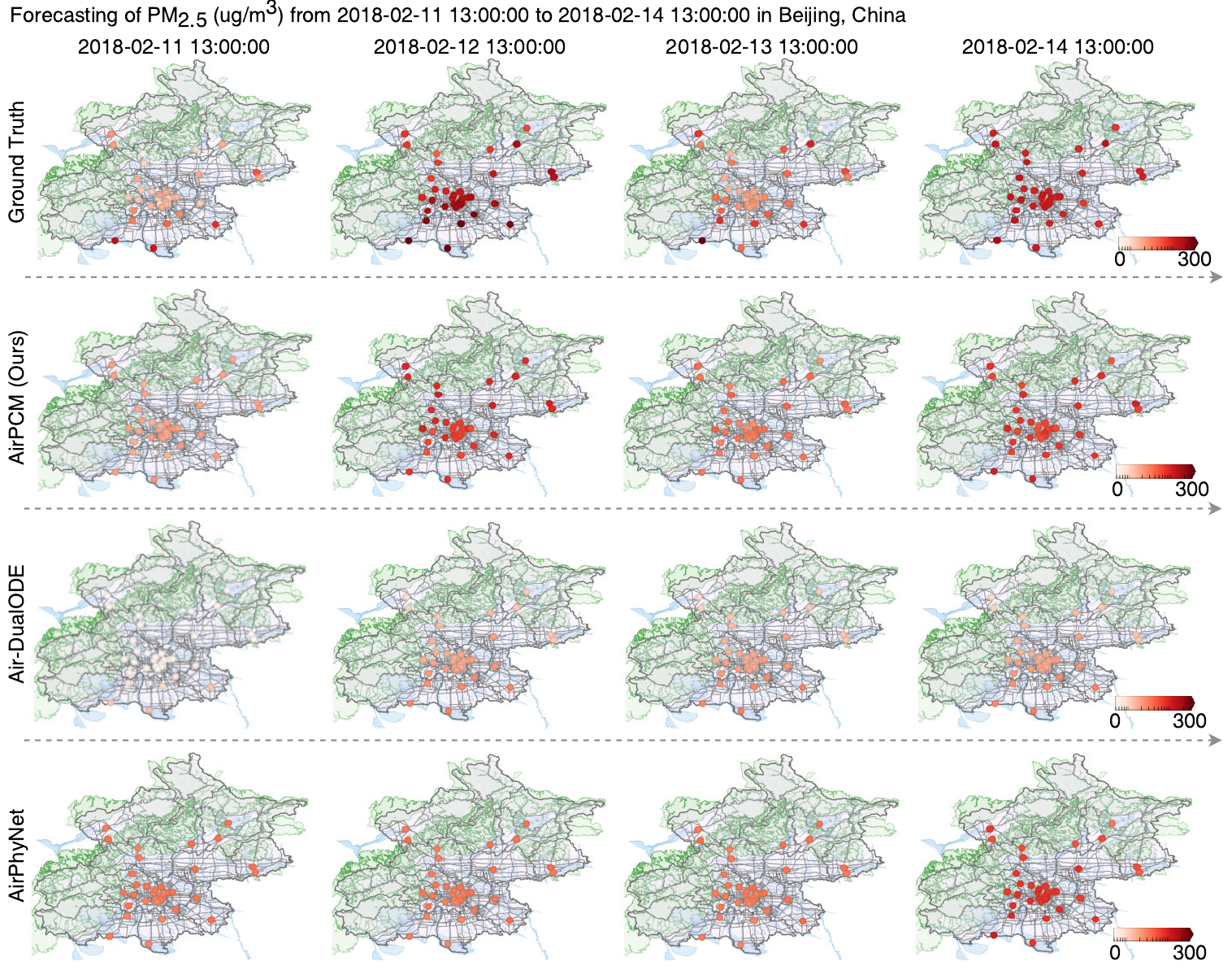}  
\caption{\textbf{PM$_{2.5}$ concentration forecasting in Beijing, China.}
On the Beijing \cite{10.1145/3219819} benchmark dataset, 
we evaluate the PM$_{2.5}$ forecasting performance of AirPCM 
against two state-of-the-art methods, Air-DualODE \cite{tian2025air}
 and AirPhyNet \cite{hettige2024airphynet}. 
Each model uses as input three days of historical observations 
(from 13:00 on February 8 to 13:00 on February 11, 2018) 
to forecast PM$_{2.5}$ concentrations for the subsequent three-day period 
(from 13:00 on February 11 to 13:00 on February 14, 2018).
In the visualization, darker colors denote higher predicted concentrations 
at each monitoring station.
}
\label{beijing_pm25_Result}
\end{figure*}

\begin{figure*} 
\centering
\includegraphics[width=0.99\textwidth]{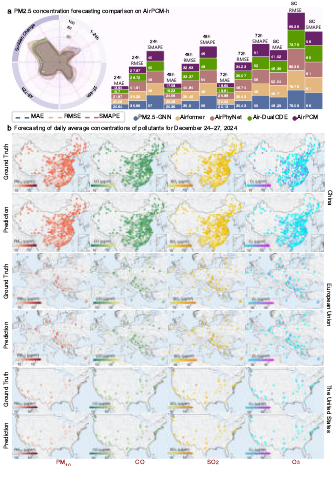}  
\caption{\textbf{AirPCM forecasting performance on AirPCM-h dataset.}
\textbf{a,} Predicted PM2.5 concentrations for the future three days 
by AirPCM on the self-collected hourly dataset AirPCM-h 
(smaller areas indicate lower prediction errors and thus  higher accuracy).
\textbf{b,} Daily average pollutant concentrations from December 24 to 27, 2024, 
 comparing AirPCM predictions with actual observations on the AirPCM-h dataset.
}
\label{mydata_results}
\end{figure*}

\textbf{Qualitative comparison.} 
Fig. \ref{beijing_pm25_Result} qualitatively compares PM$_{2.5}$ forecasts in Beijing 
 produced by AirPCM and two leading baselines, Air-DualODE \cite{tian2025air} and AirPhyNet \cite{hettige2024airphynet}, 
over a three-day horizon starting on February 11, 2018. 
Overall, AirPCM yields predictions that align more closely with the observed concentrations, 
exhibiting smaller deviations throughout the forecast horizon. 
Although all models display some lag or underestimation during abrupt changes, 
AirPCM demonstrates more stable tracking of pollution levels across 
varying temporal segments.

To further evaluate the performance of AirPCM on multi-region 
and multi-pollutant forecasting, we collected a comprehensive dataset, AirPCM-h, covering three representative 
regions: the European Union, the United States, and China. 
The AirPCM-h dataset contains hourly air pollution and meteorological observations
from January 17, 2024, to April 17, 2025, across a total of 453 monitoring stations,
with 55 located in Europe, 35 in the United States, and 363 in China.
The spatial distribution of these stations is shown in Fig. \ref{mydata}d.

Fig. \ref{mydata_results}a shows a comparative analysis of AirPCM against
several state-of-the-art baseline methods for PM$_{2.5}$ prediction on the AirPCM-h dataset. 
The results demonstrate that AirPCM consistently surpasses other mainstream models
 across both short-term (1-24 hours) and mid-term (25-72 hours) forecasting horizons.
These findings underscore the strong adaptability 
and performance of AirPCM for multi-temporal and multi-regional air quality forecasting tasks.

In addition, Fig.~\ref{mydata_results}b presents the prediction results of AirPCM for pollutants 
including PM$_{10}$, CO, SO$_2$, and O$_3$ during the period from December 24 to 27, 2024. The predicted 
daily average concentrations closely match the observed values, demonstrating the strong generalization 
capability of AirPCM and its potential for joint multi-pollutant forecasting.

\begin{figure*} 
\centering
\includegraphics[width=0.85\textwidth]{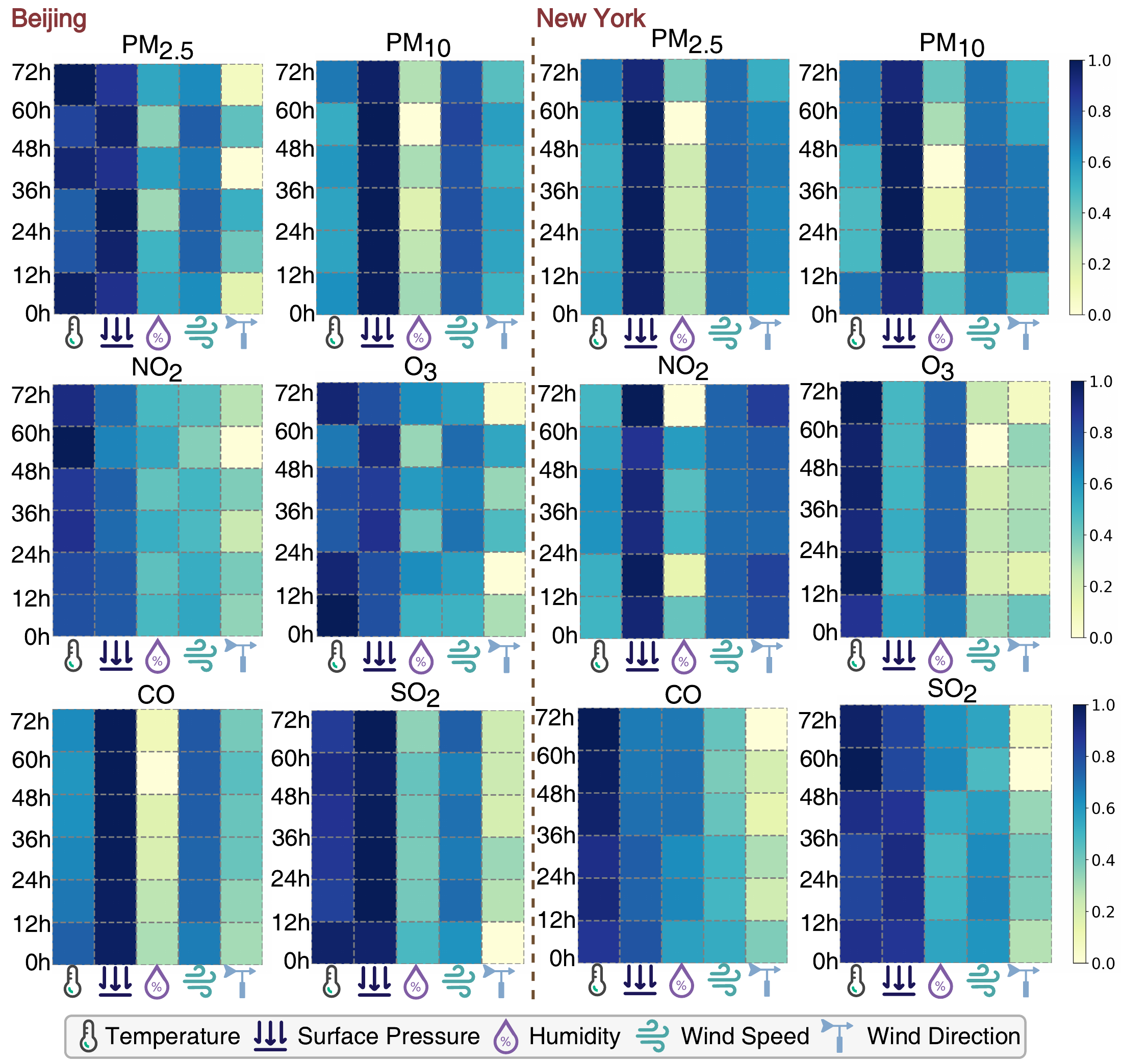}  
\caption{\textbf{Visualization of meteorology-pollutant temporal causality.}
Illustration of the causal effects of meteorological variables on the concentrations of major 
air pollutants in Beijing and New York during the period from 12:00 on January 8, 2025, to 12:00 on January 11, 2025.}
\label{causal_fig}
\end{figure*}

\subsection{Meteorology-pollutant temporal causality} 
We further analyze the dynamic temporal causal effects of meteorological variables on various air 
pollutants. As shown in Fig. \ref{causal_fig}, 
we visualize the temporal causal effects over a 72-hour horizon between five meteorological 
variables
(temperature, surface pressure, humidity, wind speed, and wind direction) and six major air pollutants 
(PM$_{2.5}$, PM$_{10}$, O$_3$, NO$_2$, SO$_2$, and CO) in two representative cities Beijing and New York.
The results reveal substantial variations in both the temporal magnitude and 
structure of the causal effects,
 with pronounced time-delay patterns and city-specific effect signatures.

Overall, the temperature exhibits a stable and pronounced positive causal effect on O$_3$ concentrations, 
particularly within short-term windows (0-24 h), aligning with the established physical mechanism whereby elevated temperatures accelerate photochemical reactions.
This observation  validates  the capacity of AirPCM to capture temperature-driven ozone dynamics
Furthermore, a comparative analysis between these two cities reveals meaningful differences, i.e., 
while the temperature-O$_3$ causality response in Beijing is sharper and more concentrated within 
a narrow time  window, whereas in New York it extends beyond 48 hours,
reflecting differences in the duration and intensity of photochemical processes 
under distinct  climatic backgrounds. 
Similarly, the humidity and the wind speed generally exert negative causal effects on 
particulate matter (PM$_{2.5}$ and PM$_{10}$), underscoring their importance in particle dispersion and deposition.
However, these effects manifest differently, Beijing shows a more immediate and steep response,
 peaking within 24-36 hours, whereas New York displays a more gradual decay over an extended period, 
 likely due to the influence of regional transport and evolving air mass dynamics.

Although the surface pressure and the wind direction exhibit relatively moderate 
causal contributions,
they still reveal consistent delayed effects for specific pollutants. 
For instance, in Beijing, the surface pressure exerts a mild positive impact on NO$_2$ and SO$_2$ 
at longer time lags, while in New York, the wind direction becomes increasingly influential for 
NO$_2$ and SO$_2$
during the 60-72 h period. These patterns  suggest that  regional pollutant accumulation or cross-boundary transport may be facilitated under specific pressure or wind regimes.
 Notably, although such influences are less prominent in the short term, 
they provide valuable insights for medium-term to long-term pollution evolutions 
and offer potential indicators for regulatory references.

The ability to uncover these temporally causal relationships opens up new 
avenues for fine-grained modeling of pollutant formation mechanisms 
and supports the development of differentiated air quality management strategies across cities. 
From a practical perspective, our analysis suggests that Beijing is more sensitive to 
abrupt meteorological 
disturbances and therefore requires responsive mechanisms to short-term changes 
(e.g., sudden wind bursts or humidity shifts).  In contrast, the pollution dynamics in New York
are more affected by delayed meteorological influences, which makes it suitable for incorporating 
middle-term weather processes and antecedent meteorological conditions into forecasting 
and regulation efforts. Importantly, our meteorology-pollutant temporal causality modeling 
enables precise quantification of marginal effects and response windows for 
each meteorological variable for pollutants,
thereby fostering interpretable predictive models and proactive early-warning systems.

\subsection{Generalization and robustness evaluation}
\begin{figure}[t]
\centering
\includegraphics[width=0.99\textwidth]{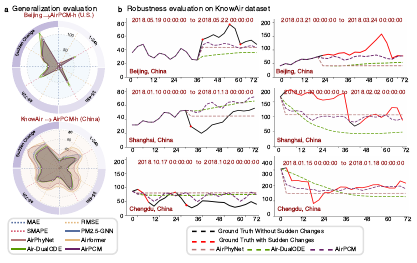}  
\caption{\textbf{Generalization and robustness evaluation.}
\textbf{a,} Generalization performance of models trained on the KnowAir and 
Beijing datasets, evaluated on the AirPCM-h dataset.
\textbf{b,} Visualization comparing model robustness in the presence of sudden changes.
}
\label{gen_and_robust}
\end{figure}

 Fig. \ref{gen_and_robust}a presents the results of generalization evaluation, 
covering two settings:
(1) models trained on the Beijing and KnowAir datasets and evaluated on the AirPCM-h dataset in the U.S. and China regions,
 respectively,
 and (2) forecasting performance assessments for PM$_{2.5}$ concentrations over the next 72 hours, 
 along with  under sudden changes.

For the KnowAir $\to$ AirPCM-h (China) transfer, the primary focus is on temporal generalization.
Here, the KnowAir dataset contains air quality data from 2015 to 2018, while AirPCM-h spans from 2024 to 2025.
Although both datasets cover China, their substantial temporal gap imposes a challenging test 
for capturing long-term temporal dynamics. 
Results in Fig. \ref{gen_and_robust}a
 demonstrate that AirPCM  consistently achieves superior performance on this task,
yielding the lowest MAE and RMSE across all horizons and sudden change cases.
This highlights its strong capability in modeling long-range temporal evolutions.

In contrast, the Beijing $\to$ AirPCM-h (U.S.) transfer primarily evaluates cross-regional generalization.
Given the stark differences in geography, climate, and pollution source distributions between Beijing and the U.S., 
this represents a demanding test of spatial adaptability.
AirPCM shows remarkable performance in this setting as well, especially under sudden changes, 
where it achieves significantly lower prediction errors than other methods, 
underlining its robust cross-regional transfer capability.

Quantitatively,  AirPCM achieves the best results across most forecasting settings.
Under sudden changes, it attains MAE, RMSE, and SMAPE of 31.89, 57.19, and 0.94 for the Beijing $\to$ AirPCM-h (U.S.) 
transfer, corresponding to gains of 3.8$\%$, 1.8$\%$, and 7.8$\%$ over the second-best method, Air-DualODE.
Similarly, for the KnowAir $\to$ AirPCM-h (China) transfer, it achieves 36.56, 54.62, and 0.51, improving by 4.0$\%$, 3.2$\%$, and 5.6$\%$ 
respectively.
These results collectively emphasize AirPCM superior generalization ability and robustness under cross-regional and temporal forecasting.

To assess the robustness of AirPCM and state-of-the-art baselines   
under sudden air quality changes, we performed experiments on the KnowAir dataset.
The evaluation focuses on PM$_{2.5}$ predictions at three representative regions in Beijing, Shanghai, 
 and Chengdu of China, as visualized in Fig.~\ref{gen_and_robust}b.
The results show that AirPCM consistently provides more accurate tracking of both rising
 and falling transitions than Air-DualODE and AirPhyNet, 
which often lag behind or underreact to sudden changes, leading to smoothed local peaks and troughs. 
In contrast, AirPCM captures sharper fluctuation patterns that remain closer to the observed ground truth. 
This suggests that achieving reliable air quality forecasts under rapidly changing conditions depends 
not only on modeling average temporal trends but also on learning to represent transient dynamics, 
including brief pollutant accumulation or dispersal events influenced by meteorological or anthropogenic factors.

This capability is crucial for real-time public health alerts and urban management, 
where anticipating imminent air quality deterioration enables timely interventions.
Nevertheless, AirPCM still faces challenges in precisely predicting the exact magnitudes at peak points,
underscoring that forecasting abrupt air quality episodes remains a demanding and important research task.

\subsection{Case study of long-term forecasting}
To comprehensively evaluate the practical utility of AirPCM, 
we conduct a case study over China using the AirPCM-d dataset. 
This dataset includes daily records of six key pollutants (PM$_{2.5}$, PM$_{10}$, O$_3$, NO$_2$, SO$_2$, and CO) 
and meteorological variables (temperature, pressure, humidity, wind speed, wind direction) collected 
from 156 monitoring stations across the country, covering the period from 2015 to 2025. 
Such extensive spatial and temporal coverage allows us to examine long-term air quality trends and generate 
forecasts for the subsequent two years, providing a  new perspective to explore both urban and regional pollution patterns.

\begin{figure}[t]
\centering
\includegraphics[width=0.99\textwidth]{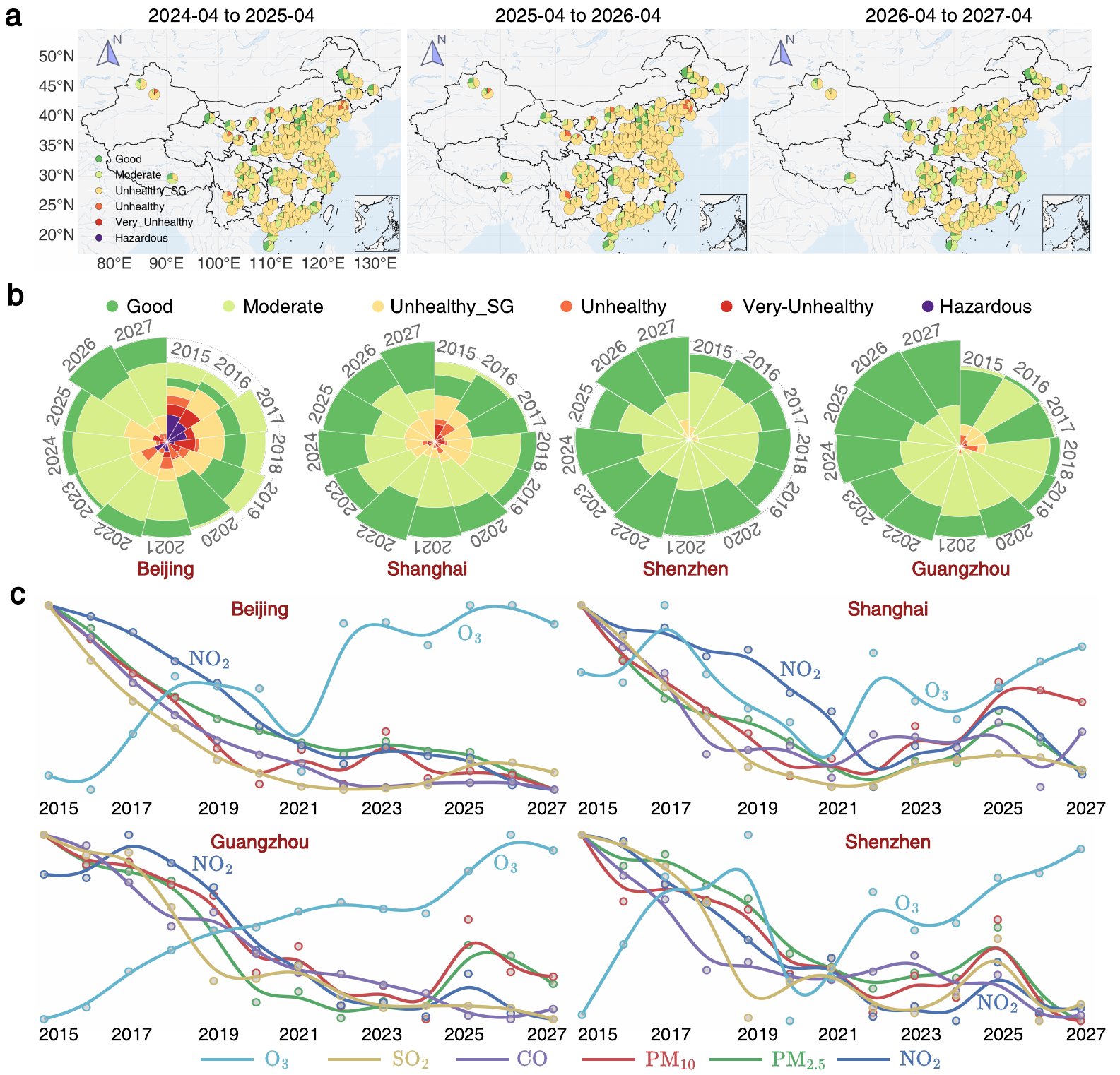}  
\caption{\textbf{Case study on future air quality index analysis in China.}
\textbf{a,} Predicted proportions of air pollution levels by AirPCM from April 2024 to April 2027 across 156 major Chinese cities, 
expressed as the percentage of days falling into each pollution category relative to the total days in a year.
\textbf{b,} Visualization of the distribution of days across different pollution 
levels for four first-tier Chinese cities from 2015 to 2027.
\textbf{c,} The trends in annual average concentrations of different 
pollutants for four first-tier cities in China from 2015 to 2027.} 
\label{AQI}
\end{figure}

As illustrated in Fig.~\ref{AQI}a, the predicted distribution of AQI categories 
(good, moderate, unhealthy for sensitive groups, unhealthy, very unhealthy, and hazardous) 
from April 2024 
to April 2027 highlights a clear north-south gradient. 
Regions in northern China, particularly Gansu, Qinghai, Liaoning, 
Neimenggu, and parts of Hebei and Shanxi, are projected to experience 
a higher fraction of days falling into 
unhealthy or worse categories. This occurs despite the implementation of 
consistent national emission standards, 
indicating that meteorological and geographical factors, 
such as frequent temperature inversions, shallow boundary layers, 
and limited atmospheric mixing over the North China Plain, play a dominant role in hindering pollutant dispersion. 
By contrast, southern and coastal regions benefit from stronger winds and more frequent precipitation events, 
which facilitate pollutant removal through both advection and wet deposition, resulting in generally better air quality.

Zooming into the four major metropolitan areas Beijing, Shanghai, Guangzhou and Shenzhen, 
Fig. \ref{AQI}b traces their historical and future air quality trajectories.
All four cities show a consistent trend toward improved air quality, marked by a steady decline in the number of days exceeding moderate AQI thresholds. 
This improvement reflects the cumulative impact of stringent emissions controls and industrial restructuring. 
In Beijing periodic severe pollution episodes occur, driven by winter coal heating and unfavorable meteorological conditions. 
Shanghai and Guangzhou follow a stable upward trend toward cleaner air though both cities continue to record occasional unhealthy days. 
Shenzhen maintains the cleanest air quality among the four thanks to coastal ventilation and proactive industrial transitions.

Fig. \ref{AQI}c further represents the annual concentration trends of six major pollutants.
It can be observed that since 2025, the concentrations of primarily pollutant 
(PM$_{2.5}$, PM$_{10}$, NO$_2$, SO$_2$ and CO)
have exhibited a sustained downward trend across the four cities, albeit with occasional fluctuations.
However, secondary pollutants, e.g., ozone (O$_3$), exhibit an upward trend.
This divergence is primarily attributed to national emission control 
and pollution mitigation policies targeting primary pollutants.
As a typical secondary pollutant, O$_3$ formation is governed by
 the complex interplay of multiple factors, 
including the concentration ratios of its precursors
 (such as volatile organic compounds VOCs and nitrogen oxides NO$_x$), 
solar radiation intensity, and meteorological conditions. 
Under current policy emphasizing NO$_x$ emission reductions, 
some regions may enter a VOC-limited regime, where decreasing 
NO$_x$ paradoxically reduces O$_3$ titration, thereby facilitating increased ozone formation.
Additionally, global warming has contributed to higher temperatures 
and enhanced solar irradiance, 
both of which accelerate photochemical reactions and further promote O$_3$ generation.

This case study demonstrates that significant spatial and 
temporal disparities in air quality
can persist under uniform regulatory frameworks, 
largely as a result of intrinsic meteorological 
and topographical factors. 
In addition, the shift in urban emission patterns,
 marked by reductions in primary pollutants and increases 
 in secondary pollutants such as ozone, 
 highlights the necessity of multi-pollutant forecasting.
It also emphasizes the importance of advanced forecasting systems like
AirPCM that integrate multi-pollutant and multi-region dynamics while explicitly modeling the causality 
  between atmospheric variables and pollution processes. By capturing multi-scale 
spatiotemporal dependencies and learning these causal influences, such systems  
are capable for proactive public health protection and the development 
of adaptive environmental policies. As urbanization accelerates and climate variability increases,
causality-informed, multi-pollutant forecasting 
frameworks like AirPCM will be critical for moving beyond reactive responses towards 
more anticipatory air quality management.

\section{Discussion}
In this study, we introduce AirPCM, a novel deep spatiotemporal model designed to address key limitations in existing air quality forecasting methods
 \cite{tian2025air,liang2023airformer,hettige2024airphynet,wang2020pm2}.
By integrating multi-region, multi-pollutant, 
and meteorological dynamics within a  causality-aware framework, 
AirPCM transcends models that narrowly focus on isolated regions and individual pollutants.
It represents a necessary shift toward capturing the interwoven dynamical dependencies that govern atmospheric pollution,
offering superior predictive accuracy and interpretability under both routine and extreme conditions.

Our proactive case study of long-term forecasting across China demonstrates that regional air pollution levels are shaped not 
only by nationwide emission-control policies but also by local meteorological conditions and cross-city pollution dispersion. 
AirPCM reveals that persistent north-south disparities in pollution are tightly coupled with regional topography and meteorological patterns, 
such as thermal inversions and prevailing winds \cite{silva2017future}. 
 This underscores a critical limitation of emission-centric strategies: 
without explicitly accounting for regional meteorological variability and cross-regional diffusion patterns, 
policy effectiveness becomes spatially heterogeneous and often suboptimal.

By jointly modeling multiple pollutants (e.g., PM$_{2.5}$, PM$_{10}$, SO$_2$, and O$_3$),
AirPCM further addresses the evolving challenge of pollutant interactions and transport.
As urban emission profiles shift—characterized by reductions in primary pollutants and increases 
in secondary pollutants like ozone—a multi-pollutant forecasting method becomes indispensable. 
Our results demonstrate that joint modeling not only improves predictive accuracy 
but also yields critical diagnostic insights into pollutant interdependencies.
This capability is crucial for devising multifaceted pollution mitigation strategies,
particularly for complex pollution events involving simultaneous surges of particulate matter and ozone.

Accurately predicting the timing and magnitude of abrupt air quality degradation remains a formidable challenge. 
While AirPCM, like other state-of-the-art models, does not fully capture the sharpest pollution spikes, 
it excels at identifying the onset and resolution phases of severe episodes. 
By closely tracking the dynamic trends before and after such events, it enables early warnings that
are instrumental in shifting from reactive responses to proactive urban management and timely public health interventions.

AirPCM transcends conventional single-pollutant, region-isolated frameworks by introducing  
a deep spatiotemporal model that simultaneously captures multi-region, multi-pollutant, and meteorological dependencies. 
Unlike previous models that overlook spatial spillovers or the lagged causal effects of meteorological factors, 
AirPCM explicitly learns these cross-scale dependencies. 
This enables it not only to predict routine pollution patterns, 
but also to more reliably signal imminent and abrupt air quality degradations.
As urbanization accelerates and climate variability intensifies, the deployment of such fine-grained and 
adaptive forecasting models is essential for effective policy-making and safeguarding of 
the public health.

\section{Methods} 
In this section, we present the methodological details of AirPCM, 
along with the datasets, baselines, and evaluation metrics used in experiments.

\subsection{Air quality forecasting based on history observations} 
Multi-region, multi-pollutant air quality forecasting relies on historical observation data from 
$N$ geographically distributed stations.  
Each station records concentrations of $K$ air pollutants (e.g., PM$_{2.5}$, PM$_{10}$ and SO${_2}$) 
along with  $C$ meteorological variables.  
This historical data form  a continuous multivariate time series, represented as $\{(x_1,m_1),(x_2,m_2),\cdots,(x_t,m_t),\cdots\}$,
 where $x_t \in \mathbb{R}^{N\times K}$ and  $m_t\in \mathbb{R}^{N\times C}$ represent the pollutant and meteorological observations
 for all stations at time $t$, respectively.
Our objective  is to forecast  future $\kappa$ time steps  pollutant concentrations $x_{t+1:t+\kappa} \in \mathbb{R}^{N \times K \times \kappa}$ 
for all stations based on past $\tau$ time steps of observations $x_{t-\tau:t} \in \mathbb{R}^{N \times K \times \tau}$ and 
$m_{t-\tau:t} \in \mathbb{R}^{N \times C \times \tau}$. Currently, under a data-driven paradigm, 
leveraging partial historical observations from geographically  scattered stations to achieve continuous, 
fine-grained, and multi-pollutant forecasting remains an 
unexplored and challenging task. 
To promote the resolution of this task, we propose AirPCM, a deep spatiotemporal model that jointly captures cross-station spatial correlations, 
temporal auto-correlations,
 and the dynamic temporal causality between meteorological variables and pollutants.

\subsection{AirPCM}
\textbf{Architecture overview.} AirPCM consists of a four-stage pipeline: multi-station spatial correlation modeling (MSCM), 
followed by patching and embedding (P$\&$E), meteorology-pollutant temporal causality learning (MPTC) and finally decoding (DECO).
MSCM receives historical observations of both pollutants and meteorological variables to learn cross-station spatial correlations, 
capturing the spatial dependencies among different stations at each time step. 
Based on these learned correlations, pollutant propagation is performed across stations to model spatial interactions. 
Following MSCM, the model performs patching and embedding on the pollutant features of each station. 
Specifically, the features are divided into equally sized patches, 
which are then embedded to project them into a latent space for forecast modeling in subsequently stages.
The four-stage pipeline of AirPCM is illustrated in Fig. \ref{overview}(b).

\textbf{MSCM.} Based on the geographical information of stations, including longitude, latitude, and altitude, 
we construct a geospatial graph $\mathcal{G}=\{\mathcal{V},\mathcal{E}\}$, where $\mathcal{V}$ is the set of stations with $|\mathcal{V}|=N$, 
and $\mathcal{E}$ denotes edges expressing potential pathways for inter-station pollutant transport  \cite{tian2025air,wang2020pm2}.
Given the historical pollutant observations $x_{t-\tau:t} \in \mathbb{R}^{N \times K \times \tau}$,  
we first apply a two-dimensional convolution (Conv) to project inputs into a latent feature space while capturing local spatiotemporal patterns, i.e.,
$h_{t-\tau:t}=\text{Conv}(x_{t-\tau:t}) \in \mathbb{R}^{N \times d_h \times \tau}$, where $d_h$  is the hidden dimension.
Subsequently, we employ  a graph attention network (GAT) \cite{velivckovic2018graph}   over  $\mathcal{G}$ 
to aggregate spatial correlations across 
stations, yielding node geospatial representations $\tilde{h}_{t-\tau:t}=\text{GAT}(h_{t-\tau:t}, \mathcal{G}) 
\in \mathbb{R}^{N \times d_h \times \tau}$.
To further capture non-local dependencies across stations, we incorporate a multi-head self-attention (MSA) mechanism \cite{vaswani2017attention} with 
queries $q=W_qh$ from $h_{t-\tau:t}$  and key-value pairs $k=W_k\tilde{h}$, $v=W_v\tilde{h}$ from $\tilde{h}_{t-\tau:t}$:
\begin{equation}
  \text{MSA}(q,k,v)=\text{Softmax}(\frac{qk^{\top}}{\sqrt{d_{msa}}})v,
\end{equation}
where $d_{msa}$  is the per-head dimension.  
The attended outputs are passed through residual connections
and a position-wise feed-forward network with layer normalization (FFN). Finally, a convolution projects 
the features back to the original observation space, 
yielding spatially contextualized observations for downstream temporal forecasting. This process can be described as:
\begin{equation}
  x'_{t-\tau:t} = \text{Conv}\left(\text{FFN}\left(x_{t-\tau:t}+\text{MSA}(q,k,v)\right) \right)  \in \mathbb{R}^{N \times K \times \tau}.
\end{equation} 
 
\textbf{P$\&$E.} We apply patching along the temporal dimension for each pollutant variable, generating overlapping patches.
Take PM$_{2.5}$ as an example. The observations $x'^{\text{PM}_{2.5}}_{t-\tau,t} \in \mathbb{R}^{N \times \tau}$ are
segmented  into patches $p \in \mathbb{R}^{N \times n_p \times l_p}$, where $l_p$ is the patch length and
$n_p=\left\lfloor \frac{(\tau - l_p)}{s}\right\rfloor+2$ is the total number of patches given a sliding stride $s$.
We then employ embedding layers to project each patch into a latent space while incorporating its positional and temporal attributes.
Specifically, for each patch, we aggregate  its position $p_{pos} \in \mathbb{R}^{N \times n_p \times 1}$ 
and temporal information $p_{time} \in \mathbb{R}^{N \times n_p \times 4}$ (start year, month, day, and hour) as follows:
\begin{equation}
  p_{emb} = \text{Embed}_{patch}(p) + \text{Embed}_{pos}(p_{pos}) + \text{Embed}_{time}(p_{time}), 
\end{equation} 
where $p_{emb} \in \mathbb{R}^{N \times n_p \times d_p}$ is the embedded patch with embedding dimension $d_p$.  
Note that the embedding layer $\text{Embed}_*(\cdot)$  adopt distinct linear projection dimensions to ensure dimensional 
compatibility across patch content, position, and temporal encodings \cite{pan2024s,nietime,wu2023interpretable}.

\textbf{MPTC.} This stage is designed to capture the temporal causal effects of meteorological
 variables on pollutant evolution.
Concretely, given meteorological observations $m_{t-\tau:t} \in \mathbb{R}^{N \times C \times \tau}$, 
they are first processed by the MSCM module, yielding spatially contextualized features 
$m'_{t-\tau:t} \in \mathbb{R}^{N \times C \times \tau}$ that encode cross-station dependencies.  
Then, we extract the most recent $\omega$ time steps 
(time-delayed causal window \cite{lu2024caudits,sasaki2023dryland,fu2024generating}) of 
meteorological observations $m'_{t-\omega:t} \in \mathbb{R}^{N \times C \times \omega}$ 
for time step $t$, which are then projected via an embedding layer into a latent space,
 resulting in meteorological embeddings 
$p_{met} = \text{Embed}_{met}(m'_{t-\omega:t}) 
 \in \mathbb{R}^{N \times C \times \omega \times d_p}$. 
 To explicitly model the causal influence of meteorology on pollutants, 
 we employ a multi-head causal attention mechanism (MCAM)
  \cite{vaswani2017attention}. Here, the query representations are obtained from the pollutant patch embeddings $p_{emb}$,
  while the keys and values are derived from meteorological embeddings $p_{met}$. 
Meanwhile, to preserve rational temporal causality,  a lower-triangular mask $M \in \{0,1\}^{n_p \times C \times \omega}$  
 is imposed over the attention scores. 
This is formalized as: 
\begin{equation}
p_{mcam} = \text{MCAM}(\hat{q},\hat{k},\hat{v})=\text{Softmax}(\frac{\hat{q}\hat{k}^{\top}}{\sqrt{d_{mcam}}}\odot M)\hat{v},
\end{equation}
where $\hat{q}=\hat{W}_qp_{emb}$, $\hat{k}=\hat{W}_kp_{met}$, and $\hat{v}=\hat{W}_vp_{met}$. 
This step ensures that each pollutant query patch selectively attends only to historical meteorological features within its causal window,
 yielding interpretable temporal influence patterns.

\textbf{DECO.}  The causally attended outputs $p_{mcam}$ are first fused with the original 
pollutant patch embeddings $p_{emb}$ to obtain the initial input 
$x_{deco}^{0}=W_f[p_{mcam}, p_{emd}]  \in \mathbb{R}^{N \times n_p \times d_p}$ for decoding. 
The representations are then iteratively updated by a stack of temporal self-attention blocks, 
each consisting of an MSA followed by an FFN with residual connection:
\[
\tilde{x}_{deco}^{l-1} = \text{MSA}\bigl(x_{deco}^{l-1}\bigr), 
x_{deco}^l  = \text{FFN}\bigl(\tilde{x}_{deco}^{l-1}\bigr),
\quad l=1,\dots,L,
\]
where $x_{deco}^l \in \mathbb{R}^{N \times n_p \times d_p}$ denotes the output of the $l$-th  block 
 and $L$ is the total number of stacked temporal self-attention blocks.

To accommodate the heterogeneous dynamics across different pollutant targets, 
we further introduce pollutant-specific adapters that tailor the decoded features to each target variable. 
The adapted features are flattened and fed into dedicated output projectors to produce multi-horizon forecasts:
\begin{equation}
  \hat{x}_{t+1:t+\kappa} = \mathcal{O}_k (\text{Flatten}(\text{Adapter}_k(x_{deco}^{l})))  \in \mathbb{R}^{N \times K \times \kappa}, 
\end{equation}
where $\hat{x}_{t+1:t+\kappa}$ is  the predicted pollutant concentrations over the future $\kappa$ time steps
for all $K$ target pollutants across $N$ stations. Here,  $\text{Adapter}_k(\cdot)$ and $\mathcal{O}_k(\cdot)$ 
denote the target-specific adapter and output projector for pollutant $k$, respectively. 

\begin{figure}[H]
\centering
\includegraphics[width=0.95\textwidth]{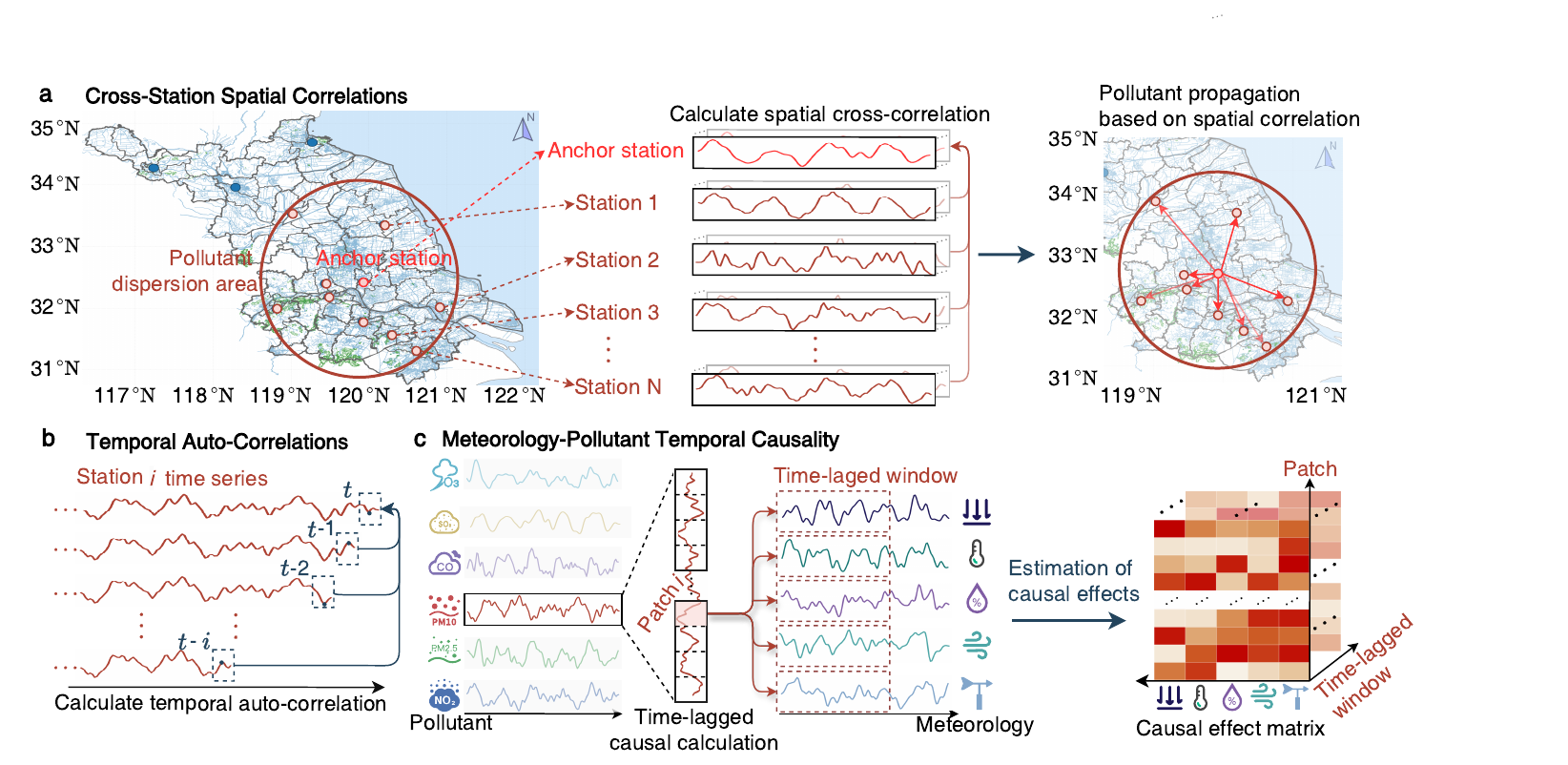}  
\caption{\textbf{Overview of modeling spatiotemporal-causal relationships in AirPCM.} 
\textbf{a,} Cross-station spatial correlations. 
\textbf{b,} Temporal auto-correlations. 
\textbf{c,} Temporal causality between meteorological variables and pollutants.}
\label{space_time_causal}
\end{figure}
 
\subsection{Summary of AirPCM} 
AirPCM is a unified deep spatiotemporal model designed to capture complex dependencies for multi-region,
 multi-pollutant air quality forecasting. As illustrated in Fig.~\ref{space_time_causal}, it 
 systematically integrates three major dependency modeling independents which are highlight in below.
 
 \textbf{(i)  Cross-station spatial correlations.}
Rather than relying on brute-force pairwise computations, AirPCM construct a geospatial graph over all stations and applies 
 a graph attention network (GAT) to dynamically aggregate information from geographically correlated neighbors. 
 By integrating these GAT-refined features with a multi-head spatial self-attention layer, 
 AirPCM effectively captures both local and global spatial dependencies among scattered stations with linear complexity.

\textbf{(ii)  Temporal auto-correlations.}  
To capture intrinsic sequential patterns and periodic behaviors in pollutant series, 
historical observations are segmented into overlapping temporal patches and embedded into latent representations. 
These are refined via multi-head temporal self-attention blocks, 
facilitating the modeling of multi-scale temporal dependencies critical for accurate temporal forecasting.

 \textbf{(iii)  Meteorology-pollutant temporal causality.}
To  model how historical meteorological conditions drive pollutant evolution, 
we introduce a fine-grained causal attention module equipped with lower-triangular masks that strictly 
enforce temporal directionality. This design not only ensures temporal consistency 
but also yields interpretable attention weights that reveal delayed meteorological effects on pollutants.

Formally, the complete pipeline of AirPCM can be succinctly summarized as:
\begin{equation}
\scriptsize
\begin{tikzcd}[ampersand replacement= ,  ]
\text{MSCM}(x_{t-\tau:t}, \mathcal{G})
\overset{\textbf{(i)}}{\underset{x'_{t-\tau:t}}{\longrightarrow}} 
\text{P\&E}(x'_{t-\tau:t})
\overset{\textbf{(ii)}}{\underset{p_{emb}}{\longrightarrow}} 
\text{MPTC}(p_{emb}, p_{met})
\overset{\textbf{(iii)}}{\underset{p_{mcam}}{\longrightarrow}} 
\text{DECO}(p_{mcam}) \overset{\textbf{(ii)}}{\longrightarrow}
\hat{x}_{t+1:t+\kappa} 
\\
\text{MSCM}(m_{t-\tau:t}, \mathcal{G})
\overset{\textbf{(i)}}{\underset{m'_{t-\tau:t}}{\longrightarrow}} 
\text{Embed}_{met}(m'_{t-\omega:t}), \quad \quad \quad \quad \quad \quad \quad \quad \quad \quad 
\arrow[u, "p_{met}"{right},"\textbf{(ii)}"{left}]
\end{tikzcd} 
\end{equation}
where each $\overset{*}{\underset{*}{\longrightarrow}}$ denotes both the 
type of dependency being modeled (above or left of the arrow) and 
the resulting intermediate representation (below or right of the arrow).  
AirPCM jointly encodes cross-station spatial correlations, 
meteorology-pollutant temporal causality, and temporal auto-correlations, 
offering a holistic solution for fine-grained, multi-horizon air quality forecasting. 
 
\subsection{Datasets} \label{dataset}
For the air quality forecasting task, we conducted experiments  using four benchmark datasets spanning three spatial scales: 
 city-scale (Beijing \cite{10.1145/3219819} dataset),  national-scale (KnowAir \cite{wang2020pm2} 
and AirPCM-d datasets), and global-scale (AirPCM-h dataset). 
The spatial distribution of monitoring stations is illustrated in Fig. \ref{mydata}.
These datasets provide comprehensive coverage of air pollutant concentrations 
and meteorological variables, enabling a well-rounded evaluation of forecasting models 
across diverse geographical and climatic contexts. 
 We split these datasets into training, validation, and test sets using a 2:1:1 ratio.

The Beijing dataset  \cite{10.1145/3219819} provides hourly measurements 
from 35 monitoring stations in Beijing, China,  spanning 1 January 2017 to 30 May 2018. 
 It includes concentrations of six major air pollutants (PM$_{2.5}$, PM$_{10}$, O$_3$, NO$_2$, SO$_2$, and CO) 
 and meteorological variables (temperature, pressure, humidity, wind speed, and wind direction).

 The national-scale KnowAir dataset \cite{wang2020pm2} focuses on PM$_{2.5}$ forecasting across 184 cities in China.
It contains PM${2.5}$ concentrations and 17 meteorological variables 
(including temperature, pressure, humidity, wind components, and precipitation)
 recorded every three hours from 1 January 2015 to 31 December 2018.

 AirPCM-d and AirPCM-h are our curated daily and hourly datasets, respectively. 
Two datasets both contains the six major pollutants (PM$_{2.5}$, PM$_{10}$, O$_3$, NO$_2$, SO$_2$, and CO), 
Air Quality Index (AQI) calculated according
 to  MEP-2012 standard of China \cite{mep2012aqi}, and the meteorological variables (temperature, pressure, humidity, wind speed, and wind direction). 
 Specifically, the AirPCM-d dataset provides daily records of air pollutants 
and meteorological variables from 156 monitoring stations across China, covering the period from 2 January 2015 to 12 April 2025. 
This dataset supports long-term air quality forecasting at a national scale.
The AirPCM-h dataset, contains  
hourly measurements of air pollutants   and meteorological variables from 453 monitoring stations worldwide, 
including 55 in Europe, 35 in the United States, and 363 in China, spanning from 17 January 2024 to 17 April 2025. 
This dataset enables global-scale multi-pollutant air quality forecasting across diverse climatic regions.


\subsection{Baselines and evaluation metrics}   \label{baseline_and_em}
We evaluate AirPCM against a comprehensive set of baseline models spanning four major categories:

(i) Classic statistical methods: historical average (HA) \cite{zhang2017deep} and vector auto-regression(VAR) \cite{toda1993vector}; 

(ii) Differential equation network-based models: ODE-RNN \cite{rubanova2019latent}, 
Latent-ODE \cite{chen2018neural}, and  ODE-LSTM  \cite{lechner2020learning};

(iii) Spatiotemporal deep learning methods:  DCRNN \cite{li2018diffusion}, STGCN \cite{yu2018spatio}, ASTGCN \cite{guo2019attention}, GTS \cite{shang2021discrete},
 MTSF-DG \cite{zhao2023multiple}, PM$_{2.5}$-GNN \cite{wang2020pm2}, and Airformer \cite{liang2023airformer};

 (iv) Physics-guided neural model: AirPhyNet \cite{hettige2024airphynet} and Air-DualODE \cite{tian2025air}. 

All method are evaluated by mean absolute error (MAE), root mean square error (RMSE), 
and symmetric mean absolute percentage error (SMAPE), where a lower error value  indicates a superior predictive performance.
These metrics are defined as follows:
\begin{equation}
 \text{MAE} = \frac{1}{\kappa}\sum_{t+1}^{t+\kappa}\left\lvert \hat{x}_t  - x_t\right\rvert,  
\end{equation}
\begin{equation}
 \text{RMSE} = \sqrt{\frac{1}{\kappa}\sum_{t+1}^{t+\kappa} \left(\hat{x}_t  - x_t\right)^2},  
\end{equation}
\begin{equation}
 \text{SMAPE} = \frac{100\%}{\kappa}  \sum_{t+1}^{t+\kappa}  \frac{\left\lvert \hat{x}_t  - x_t\right\rvert }
 {\left(\left\lvert\hat{x}_t \right\rvert +\left\lvert  x_t \right\rvert \right)/2},
\end{equation}
where $\hat{x}_t$ is the predicted value corresponding to the ground truth $x_t$, and $\kappa$ represents the length of the time series.
The differences among these equations reflect the distinct design philosophies underlying the metrics.
MAE quantifies the average magnitude of prediction errors, serving as a direct indicator of typical forecast deviations irrespective of directions. 
RMSE emphasizes larger errors through its quadratic formulation, thereby rendering it more sensitive to outliers and highlighting 
the capability of the model to mitigate extreme discrepancies. 
SMAPE offers a scale-independent evaluation by representing errors as 
a percentage relative to both observed and predicted values, which facilitates consistent assessments across pollutants
with heterogeneous concentration ranges.


\section{Data availability}
The Beijing dataset is publicly available at \url{https://www.biendata.xyz/competition/kdd_2018/}. 
The KnowAir dataset can be accessed at \url{https://github.com/shuowang-ai/PM2.5-GNN}. 
The AirPCM-d and AirPCM-h datasets, collected by the authors, 
are available at \url{https://huggingface.co/datasets/junxinlu/AirPCM-d_and_AirPCM-h}. 
The air quality data in AirPCM-d and AirPCM-h are derived from national air quality monitoring platforms in various regions 
(China, the United States, and Europe), 
while the meteorological data are obtained from the fifth-generation European Centre for Medium-Range Weather Forecasts (ECMWF) 
atmospheric reanalysis dataset (ERA5), available at \url{https://cds.climate.copernicus.eu/datasets/reanalysis-era5-pressure-levels}.

\section{Code availability}
 Relevant code and models are available via GitHub at \url{https://github.com/JunxinLu001/AirPCM}.

\bibliography{sn-bibliography}

\section{Author contributions}
Junxin Lu conceptualized the study and performed the methodology development, experiments, and data analysis. 
Shiliang Sun supervised the study, provided critical insights, and guided the research direction. 
Junxin Lu drafted the manuscript, and both Junxin Lu and Shiliang Sun contributed to manuscript writing and revision.

\section{Competing interests}
The authors declare no competing interests.

\section{ Additional information}
\textbf{Correspondence} and requests for materials should be addressed to Shiliang Sun.

\begin{appendices}



\end{appendices}

\end{document}